\documentclass[runningheads]{llncs}
\usepackage[T1]{fontenc}

\usepackage{graphicx}
\usepackage{amsmath}
\usepackage{booktabs}
\usepackage{hyperref}
\usepackage{cleveref}
\usepackage{tikz}
\usepackage{color}

\begin{document}

\title{FETCH: A Memory-Efficient Replay Approach for Continual Learning in Image Classification}
\titlerunning{FETCH}
\author{Markus Weißflog\inst{1}\orcidID{0009-0003-1163-8755} \and
Peter Protzel\inst{1}\orcidID{0000-0002-3870-7429} \and
Peer Neubert\inst{2}\orcidID{0000-0002-7312-9935}%
}
\authorrunning{M. Weißflog et al.}
\institute{Faculty of Electrical Engineering and Information Technology,\\
Chemnitz University of Technology, Germany \\
\email{markus.weissflog@etit.tu-chemnitz.de} \and
Institute for Computational Visualistics, University of Koblenz, Germany}
\maketitle              % typeset the header of the contribution
% Link to version of record
\begin{tikzpicture}[remember picture,overlay]
    \node[anchor=north, yshift=-1cm] at (current page.north) {
      \parbox{\textwidth}{\scriptsize This preprint has not undergone peer review or any post-submission improvements or corrections. The Version of Record of this contribution is published in \textit{Intelligent Data Engineering and Automated Learning – IDEAL 2023}, and is available online at \url{https://doi.org/10.1007/978-3-031-48232-8_38}\\\rule{\textwidth}{0.4pt}}
    };
\end{tikzpicture}
\begin{abstract}
    Class-incremental continual learning is an important area of research, as static deep learning methods fail to adapt to changing tasks and data distributions.
    In previous works, promising results were achieved using replay and compressed replay techniques.
    In the field of regular replay, GDumb \cite{Prabhu20} achieved outstanding results but requires a large amount of memory.
    This problem can be addressed by compressed replay techniques.
    The goal of this work is to evaluate compressed replay in the pipeline of GDumb.
    We propose FETCH, a two-stage compression approach. First, the samples from the continual datastream are encoded by the early layers of a pre-trained neural network. Second, the samples are compressed before being stored in the episodic memory. Following GDumb, the remaining classification head is trained from scratch using only the decompressed samples from the reply memory.
    We evaluate FETCH in different scenarios and show that this approach can increase accuracy on CIFAR10 and CIFAR100.
    In our experiments, simple compression methods (e.g., quantization of tensors) outperform deep autoencoders.
    In the future, FETCH could serve as a baseline for benchmarking compressed replay learning in constrained memory scenarios.
\end{abstract}
\begin{figure}
    \centering                                % W   S   E   N   
    \includegraphics[width=\linewidth]{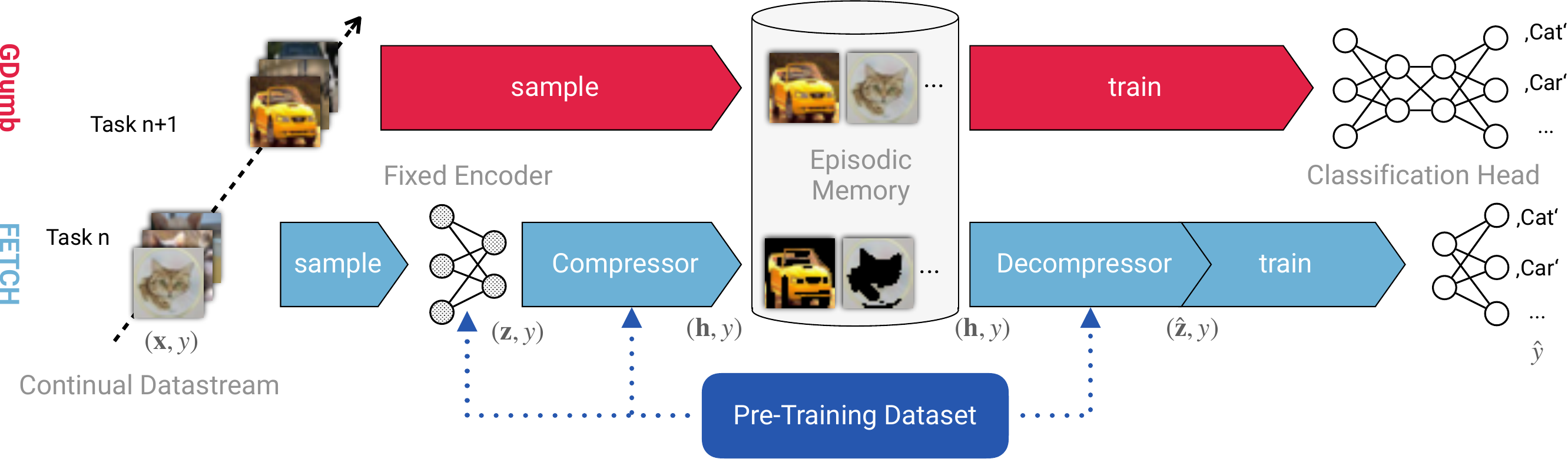}
    %\vspace{-0.75cm}
    \caption{Overview of the proposed method: GDumb \cite{Prabhu20} (red arrows) saves samples from the continual datastream in the episodic memory without preprocessing. A neural network is trained from scratch using only the samples from the memory. FETCH (blue arrows) fixes the early layers of the neural network after pre-training. The samples are saved efficiently using the compressor and decompressor. Only the classification head needs to be retrained.}
    \label{fig:intro}
\end{figure}
\section{Introduction}
\label{sec:introduction}
Humans are capable of learning to solve new tasks ever throughout their lives. Learning to incorporate new information is crucial in areas such as robotics and machine learning, yet this endeavor remains challenging despite the use of advanced techniques \cite{Parisi19}.
If no special measures are taken, a learning system quickly forgets old knowledge as soon as it is presented with new information \cite{DeLange21}. This challenge, known as Catastrophic Forgetting (CF), is so difficult to overcome that Continual Learning (CL) has emerged as a discipline of machine learning.
In recent years, many publications have approached the subject~\cite{Parisi19,DeLange21,Masana21,Mai21}. Promising results were achieved by \emph{replay techniques}, that keep an episodic memory of previously encountered samples to mitigate catastrophic forgetting \cite{DeLange21}. Prabhu et al.~\cite{Prabhu20} in particular presented \emph{GDumb}, an approach that has attracted much attention in the community due to its simplicity and yet good performance.\footnote{At the time of writing, the GDumb has received over 300 citations on Google Scholar.}
Replay techniques can have high memory consumption, which has led to the development of \emph{compressed replay methods} \cite{Hayes19,Hayes20,Wang21a,Wang22a}.
This work investigates whether the memory consumption of GDumb can be reduced using ideas from compressed replay and how the performance changes under constrained memory.
Based upon this, we present FETCH (Fixed Encoder and Trainable Classification Head). A simplified schematic illustration can be found in \cref{fig:intro}. FETCH improves GDumb in several aspects:

\begin{itemize}
    \item A pre-trained fixed encoder extracts general features from the images and enables knowledge transfer from a pre-training dataset. Additionally, the number of parameters in the trainable classification head is reduced.
    \item A compressor reduces the size of the samples in the episodic memory, thus reducing the overall memory footprint. This can either improve performance on a limited memory or reduce the memory footprint of the overall pipeline.
    \item In our experiments we assess various variations and components of FETCH and show improved performance over both GDumb and selected compressed replay techniques.
\end{itemize}

\noindent The paper is structured as follows: \Cref{sec:problem_formulation} introduces the problem of class-incremental continual learning.
\Cref{subsec:literature_review} presents general related work while \cref{subsec:gdumb} focusses on GDumb in particular.
\Cref{sec:approach} details the design of FETCH. \Cref{sec:implementation} summarizes implementation details. \Cref{sec:experiments} presents our experiments. \Cref{sec:conclusion} concludes the paper. Code will be made available.

\section{Problem Formulation}
\label{sec:problem_formulation}
Our proposed approach operates in the challenging online, class-incremental setting. A learning agent is presented with a stream of tasks $\mathcal T = \{\mathcal T^1, \mathcal T^2, \dots, \mathcal T^t,\allowbreak \dots, \mathcal T^T\}$ one task at a time. $T$ is the total number of tasks and $t$ is the current task's identifier. Each task consists of multiple samples, i.e., images, $\mathbf x^t \in \mathcal X^t$ and their corresponding labels $y^t\in \mathcal Y^t$.
The agent's job is to find a model $f$ that can predict the label for each sample $f_t: \mathbf x^t_\mathrm{test} \mapsto y^t_\mathrm{test}$, where the samples $\mathbf x^t_\mathrm{test}$ belong to a never before seen test dataset $\mathcal X^t_{\mathrm{test}}$.
The labels of these samples consist of the classes of the current task and all previous tasks $y^t_\mathrm{test} \in \mathcal Y^t_{\mathrm{test}} = \bigcup _{i=0}^t \mathcal Y^i$.
To achieve this, the agent is presented with a set of training examples belonging to the current task $\mathcal T^t = (\mathcal X^t, \mathcal Y^t)$. The agent has the option of saving a pair of image and label, or choosing to never see it again.

\section{Related Work}
\label{sec:related_work}
\subsection{Literature Review}
\label{subsec:literature_review}

Several publications offer an overview of CL. \cite{Parisi19,DeLange21,Masana21}. De Lange et al. \cite{DeLange21} in particular propose a widely adopted taxonomy, categorizing approaches and settings into parameter isolation, regularization, and replay methods.
\emph{Parameter isolation methods} work by identifying important parts of the model for the different tasks. 
These parts can be, for example, gradually extended \cite{Ardywibowo22} or even exchanged \cite{Hossain22}, as new tasks arise.
\emph{Regularization methods} introduce new terms in the loss function to ensure that performance for previous tasks is not degraded \cite{Bhat22,Collins19,Guo22}.

\emph{Replay methods} work by storing exemplars in an episodic memory and interweaving them into the stream of new data \cite{Robins95}.
Sangermano et al. \cite{Sangermano22} use dataset distillation in the replay memory. 
Chen et al. \cite{Chen22} use a database of unlabeled data together with an episodic memory to improve learning performance.
Gu et al. \cite{Gu22} propose to better utilize samples from the continual datastream to mix with the samples from the replay memory.

A special case of replay is \emph{compressed replay}. As replay requires to store a subset of exemplars in a memory, it is a sensible idea to compress these exemplars in order to reduce the overall memory footprint.
Hayes et al. \cite{Hayes19} and Wang et al. \cite{Wang22a} use different compression strategies to reduce the size of the images in memory.
Hayes et al. \cite{Hayes20} extend these approaches by freezing the early layers of a neural network after initial pre-training and using the resulting feature maps as exemplars for compressed replay. The decompressed samples are used to train the remaining parts of the network. Wang et al. \cite{Wang21a} extend this approach even further with an additional autoencoder. 
All methods operate in the classical replay scenario where the continual datastream is mixed with the stored exemplars.

\subsection {Greedy Sampler and Dumb Learner}
\label{subsec:gdumb}

GDumb (Greedy Sampler and Dumb Learner) \cite{Prabhu20} was proposed as a simple baseline but still outperformed many previous methods. It uses an episodic memory with $N$ free slots. During training, the memory slots are filled with samples from the datastream with a balancer ensuring equal class representation. When memory is full, new classes are added by removing exemplars from the largest class. After each task, GDumb retrains a backbone network from scratch using only the exemplars from the memory. Following \cite{DeLange21}, GDumb can be classified as online, class-incremental CL.
GDumb's simple design comes with some drawbacks. Saving raw data isn't always feasible due to licensing and privacy concerns. Moreover, the images require significant storage space that might not be used efficiently.
Therefore, we propose to combine the methodology of GDumb with the principles of compressed replay in order to exploit the advantages of both approaches.

\section{Approach}
\label{sec:approach}

% \begin{figure}
%     \centering
%     \includegraphics[width=\columnwidth]{images/pipeline.pdf}
%     \caption{FETCH architecture overview. The blue arrow illustrates the data flow during training, while the orange arrow represents the data flow during inference. Dotted lines represent pre-training.}
%     \label{fig:pipeline}
% \end{figure}

Following GDumb \cite{Prabhu20}, FETCH uses an episodic memory and retrains the classification head after each task. 
The blue arrows in \cref{fig:intro} show an overview of the proposed pipeline.
Whenever the input distribution changes, the data $\mathbf x$ is sampled from the continual datastream using GDumb's balanced greedy strategy. The data passes through the fixed encoder ($\mathbf z$) and the compressor ($\mathbf h$) before being stored in the episodic memory.
During the inference phase, the classification head is trained from scratch using only the decompressed data ($\hat{\mathbf z}$) and corresponding labels $y$ from the memory.
Compression and decompression are not required for inference, so the data flows directly from the encoder to the classification head.
The fixed encoder and some compressor-decompressor pairs must be pre-trained, so an additional pre-training dataset is used.

By comparing the red and blue arrows in \cref{fig:intro}, it becomes apparent that FETCH, unlike GDumb, leverages an additional fixed encoder, compressor, decompressor, and pre-training dataset. Both algorithms share the greedy sampler, memory, and retraining strategy for the classification head.
Like GDumb, FETCH can be classified as online class-incremental CL.
The following subsections describe the components in more detail.

\subsection{Fixed Encoder \& Trainable Classification Head}

First, an encoding model, called \emph{fixed encoder} in this work, converts the image to a latent representation $\mathbf z$. The classification model uses this representation $\mathbf z$ to predict the class $\hat y$ of the input data. If the encoding was successful, all the relevant information about the class is still present.
As encoders, we utilize the early layers of a CNN, whose weights remain frozen, following prior works \cite{Hayes20,Wang21a}.
To adhere to the paradigm of CL, we use different datasets for the pre-training of the encoder (called \emph{pre-training dataset}) and the training and evaluation of FETCH.
This approach allows for transfer effects from the pre-training dataset and is computationally more efficient than GDumb, as data only passes through the encoder once instead of each epoch. Also, fewer parameters need to be updated as the encoder stays fixed.
After the initial pre-training, the encoder's weights remain frozen.
The fixed encoder is considered as the early layers of a CNN, so the \emph{trainable classification head} can be regarded as the remaining layers. The classification head is trained using only the encoded data $\mathbf z$ from memory. 

For this work, different variations of the ResNet architecture \cite{He15} are used as fixed encoders and classification heads for multiple reasons:
First, ResNets have a good performance in many classification tasks \cite{He15,Masana21}. Second, implementations and pre-trained weights are available and lastly, ResNets are used in many other publications making comparison easy and fair \cite{Prabhu20,Hayes20,Wang21a}.
If not stated otherwise, we used the layers up to \texttt{conv4\_x} \cite{He15} as the fixed encoder for our experiments. The encoder pre-trained on ImageNet1k \cite{Russakovsky15}. The weights were provided by PyTorch.\footnote{Online: \url{https://pytorch.org/vision/stable/models.html}} The remaining layers including \texttt{conv4\_x} form the classification head.  %Thus, a pair of an encoder and a classifier is formed by pre-training the early layers and keeping their weights frozen, and just training the later layers during CL.
% Using the ResNet just as a classifier without a fixed encoder is straightforward: The architecture is chosen as proposed by \cite{He15}; only the last layer has to be changed so that the output neurons correspond to the number of classes in the dataset.

\subsection{Compressor \& Decompressor}

The encoded data $\mathbf z$ in memory may contain redundancies, so a second compression stage, called the \emph{compressor}, is used to reduce their size on the actual hardware. It operates on the matrix/ tensor representation of the images or featuremaps. A \emph{decompressor} restores the original encoded representation as close as possible. We have selected the approaches listed below for our experimental setup. Each method has its own hyperparameter $k$ that controls the amount of compression.

\begin{itemize}
    \item \emph{Quantization.} Quantization describes the reduction of the tensor entries to a small number of $k_\mathrm{quant}$ discrete states, which can be represented with fewer bits than the actual values. For decompression, a lookup table is used. To get the discrete states, the pre-training dataset is analyzed. The range between the highest and lowest values in the whole pre-training dataset is split into $k_\mathrm{quant}$ equally sized intervals. For all experiments, TinyImagenet \cite{Le15} was used as a pre-training dataset.
    \item \emph{Thinning.} The basic idea of this method is to keep only the most important, i.e., the largest entries. The resulting tensor is sparse and can thus be stored more efficiently. Instead of storing all entries, only the non-zero entries are saved together with their corresponding index in the tensor. Decompression is done by setting the stored indices of the output tensor to their corresponding values. All other entries are assumed to be equal to zero. The parameter $ k_{thin} \in [0, 1] $ describes the proportion of entries that are set to zero.
    \item \emph{Autoencoding.} Convolutional autoencoders are a deep learning-based approach for dimensionality reduction \cite{Hinton06,Masci11} of images. The compressor and the decompressor are typically already part of their architecture. In this work, the compressor consists of two blocks of Conv2d layers with kernel size 3 and padding 1, followed by ReLU activation and max-pooling with kernel size 2 and a stride of 2. The decompressor consists of two blocks of transposed two-dimensional convolution with kernel size 2, stride 2, and ReLU activation. For this architecture, the parameter $k_\mathrm{ae}$ describes the number of channels in the bottleneck. The autoencoder was pre-trained on the TinyImagenet \cite{Le15} dataset.
\end{itemize}

\subsection{Calculation of the Storage Consumption}

The total amount of storage $s_{\Sigma}$ that is consumed by the whole pipeline depends on several variables:
First, $s_{\Sigma}$ depends on the storage consumption of the model $s_\mathrm{model}$, which is split into fixed encoder and classification head. We used ResNets for our experiments. As fetch operates independently from the underlying model, $s_\mathrm{model}$ was set to zero in all evaluations. The resulting findings do not change since  $s_\mathrm{model}$ is a constant value that gets added to all results. This is also in line with the literature \cite{Hayes20,Wang21a,Pelosin21}.

Second, $s_{\Sigma}$ depends on the used datatypes. In this work, we used single-precision floating point numbers, meaning $s_\mathrm{float} = 4 \, \text{bytes}$. 
We used integers of size $s_\mathrm{addr} = 2 \, \text{bytes}$ as indices for matrices and arrays, as no matrix surpassed $2^{16}$ elements in our experiments.
For raw images, we assumed RGB values with 8 bits per channel, therefore we used $s_\mathrm{uint} = 1\, \text{byte}$.

Third, $s_{\Sigma}$ depends on the other components, one of which is the episodic memory with $N$ slots. Additionally, the compressor's input influences $s_{\Sigma}$. If the full model is used as a classification head (e. g. the fixed encoder gets omitted), the input is equal to the raw images $\mathbf x$ with datatype \texttt{uint8}. When a fixed encoder is used, the compressor receives the encoded images $\mathbf z$ of type \texttt{float32} as input. Let $n$ refer to the number of elements in these tensors in the corresponding case and let $s_\mathrm{uint/float}$ be the shorthand notation for the corresponding memory requirement.

\begin{table}[t]
    \caption{Computation of the total storage requirement for different compressors}
    \label{tab:computing_s_sigma}
    \resizebox{\textwidth}{!}{
    \begin{tabular}{c|c|c}
        \toprule
        \multicolumn{1}{c}{Thinning}
        &
        \multicolumn{1}{c}{Quantization}
        &
        \multicolumn{1}{c}{Autoencoding}
        \\
        \midrule
        $\begin{array}{ll}
            s_{\Sigma} = s_\mathrm{model}\\
            + N \cdot n \cdot \left\lceil (1-k_\mathrm{thin}) \cdot s_\mathrm{uint/float}\right\rceil \\
            + N \cdot n \cdot \left\lceil (1-k_\mathrm{thin}) \cdot s_\mathrm{addr}\right\rceil
        \end{array}$
        &
        $\begin{array}{ll}
            s_{\Sigma} = s_\mathrm{model}\\
            + k_{quant} \cdot s_\mathrm{uint/float} \\
            + N \cdot \left\lceil \left\lceil \log_2 k_\mathrm{quant}\right\rceil \cdot n \cdot \frac 18 \ \text{bytes} \right\rceil
        \end{array}$
        &
        $\begin{array}{ll}
            s_{\Sigma} = s_\mathrm{model} \\
            + s_\mathrm{ae} \\
            + N \cdot k_\mathrm{ae} \cdot n_{\mathbf h} \cdot s_\mathrm{float}
        \end{array}$
        \\
        \addlinespace[0.1cm]
        \bottomrule
    \end{tabular} 
    }      
\end{table}

Lastly, the total memory consumption depends on the compressor. The level of compression and, thus, the memory consumption of one exemplar can be adjusted using the compression parameter $k$. The resulting equation for the total memory requirement can be found in \cref{tab:computing_s_sigma}. For the autoencoder, $n_{\mathbf h}$ describes the number of elements in the spatial dimensions in the bottleneck and $s_\mathrm{ae}$ is the memory consumption of the autoencoder network, which varies between 4.6~KiB and 21.62~KiB, depending on the parameter $k_{\mathrm{ae}}$.

% Some exemplary values are given in \cref{tab:autoencoder_sizes}.
% 
% \begin{table}[b]
%     \centering
%     \caption{Storage sizes for the autoencoder and different compression parameters.}
%     \label{tab:autoencoder_sizes}
%     \begin{tabular}{@{\hspace{1em}}c@{\hspace{2em}}|@{\hspace{2em}}c@{\hspace{2em}}c@{\hspace{2em}}c@{\hspace{2em}}c@{\hspace{2em}}c@{\hspace{1em}}}
%         \toprule
%         $k_\mathrm{ae}$ & 1 & 2 & 4 & 8 & 16 \\
%         $s_\mathrm{ae}$ [KiB]
%         & 4.63  
%         & 5.77  
%         & 8.03  
%         & 12.56 
%         & 21.62  \\
%         \bottomrule
%     \end{tabular}
% \end{table}
%
\section{Implementation Details}
\label{sec:implementation}
The implementation was based on GDumb's publicly available PyTorch-code\footnote{Online: \url{https://github.com/drimpossible/GDumb}} \cite{Prabhu20}. No hyperparameters were changed. The training was done using SGDR-optimization \cite{Loshchilov16} with a batch size of 16 and learning rates in $[0.005\dots 0.05]$. Data regularization was done using normalization and cutmix \cite{Yun19} with $p=0.5$ and $\alpha=1$. Because only the data from the episodic memory is used for training, the backbone can be trained for multiple epochs without breaking the class-incremental paradigm. The performance is measured as the average accuracy on the test set after convergence and is always measured after the last task.
We used two datasets and models for evaluation. For the CIFAR10 dataset \cite{Krizhevsky09}, ResNet-18 \cite{He15} was used as the fixed encoder and the classification head. For CIFAR100, ResNet-34 was used as the fixed encoder and the classification head. Storage consumption was measured in mebibyte\footnote{1~MiB $=\text{2}^\text{20}$ bytes $\approx \text{10}^\text{6}$ bytes}.
\section{Experiments}
\label{sec:experiments}

\subsection{Tradeoff between Storage and Performance}
\begin{figure}[t]
    \centering
    \includegraphics[width=\textwidth]{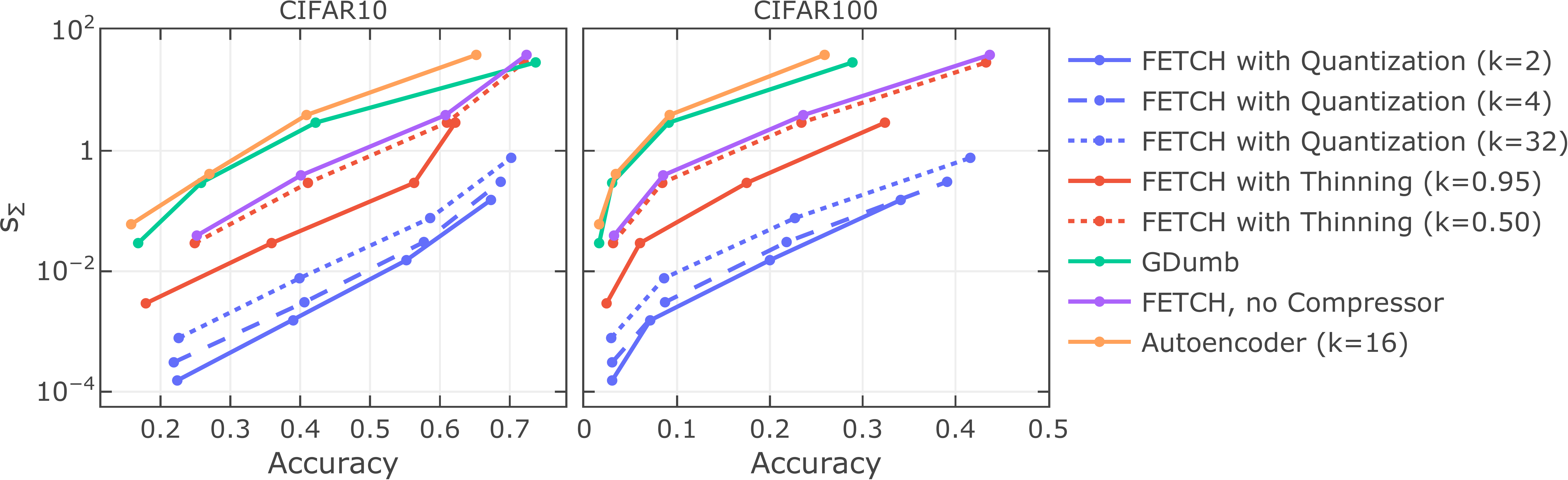}
    \vspace{-6mm}
    \caption{Tradeoff between storage and performance. Lower is better.}
    \label{fig:rescon}
    \vspace{-2mm}
\end{figure}

This experiment aimed at finding a tradeoff between low memory consumption and high accuracy. The results can be seen in \cref{fig:rescon}. The different methods were compared using a variable number of memory slots $N\in\{10; 100; 1000; 10000\}$.

%encoding improves performance?
Using a fixed encoder improves performance, as shown by the fact that most of the green curve is above the purple curves. This suggests that the transfer effects of the pre-trained fixed encoder are beneficial to the whole pipeline.
GDumb demonstrates superior performance for high storage capacities on CIFAR10, which indicates that the impact of training data quantity on performance diminishes as the learner has more memory available.
Nevertheless, model size remains a significant factor in determining performance. Restricting the number of adjustable parameters through pre-training can lead to poorer performance compared to full models.
Additionally, the quality of the data can also affect performance, as encoded samples have fewer entries and therefore provide less information to the model. For the more complex CIFAR100 and ResNet-34, this is not the case. Here, a fixed encoder is always beneficial compared to GDumb.

% compression improves performance
The impact of compression depends on the setup. Positive effects can be seen in the blue curves, which are ordered by their compression parameter $k_\mathrm{quant}$. Quantization increases performance by trading data quality for reduced storage size and thus increased storable exemplars $N$.
The red curves show a different pattern. High compression parameters, such as $k_{thin}=0.95$ show decreased storage consumption but the accuracy appears to approach an upper limit, as shown by the shape of the solid red curve in the left plot.

\subsection{Ablation: Effect of the Fixed Encoder}
\begin{figure}[t]
    \begin{minipage}[t]{0.45\linewidth}
        \vspace{5pt}
        \centering
        \includegraphics[width=0.9\linewidth]{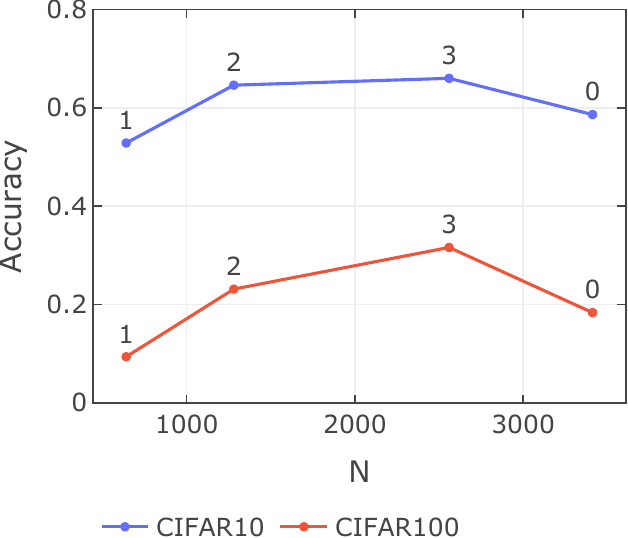}
        \vspace{10pt}
        \caption{Effect of pre-training early layers. The annotations correspond to the number of blocks in the encoder. 0 is the standard GDumb configuration, while 3 means that only the parameters of the last block were updated.}
        \label{fig:where_to_split_resnet}
    \end{minipage}\hfill
    \begin{minipage}[t]{0.49\linewidth}
        \vspace{0pt}
        \centering
        \includegraphics[width=\linewidth]{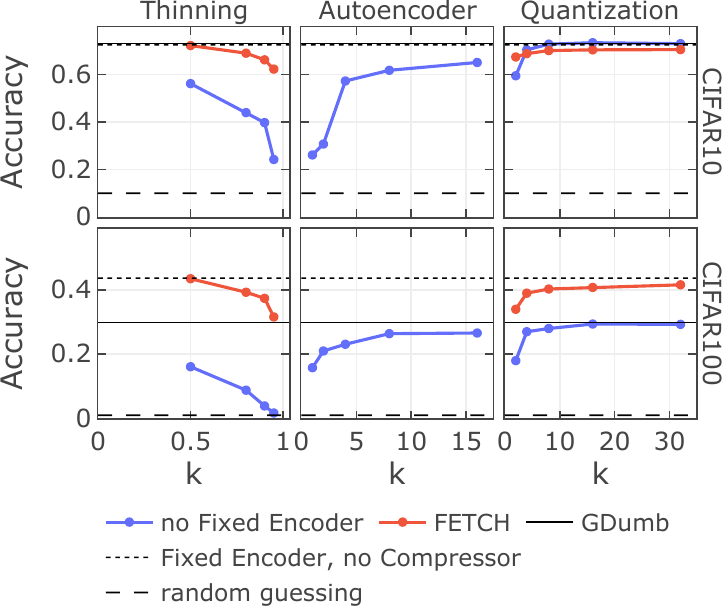}
        \vspace{-20pt}
        \caption{Effect of compressing exemplars. Note that a higher compression parameter means \textit{less} compression except for thinning, where a higher parameter means more compression. The vertical lines show the baselines without compression.}
        \label{fig:fixed_slots}
      \end{minipage}
      \vspace{-10pt}
\end{figure}

This experiment aims to investigate the influence of the layer where the ResNet is split into encoder and classification head. For this reason, the compressor was omitted. Four different configurations were investigated: Using the whole model as a classifier (like GDumb), dividing after \texttt{conv2\_x}, after \texttt{con3\_x}, and after \texttt{conv4\_x} \cite{He15}. For each configuration, the episodic memory was filled with as many samples as possible without exceeding a maximum storage of 10~MiB. 

\Cref{fig:where_to_split_resnet} shows the results. 
Splitting ResNets at later layers appears beneficial, as shown by the fact that the performance is consistently higher. At later layers, the samples are more compressed and their representation also benefits from the encoder's pre-training, which is known to be beneficial \cite{Mehta21,Pawlak22}.
Not splitting the ResNet results in the highest number of memory slots, despite raw images $\mathbf x$ having more elements than encoded samples $\mathbf z$.
Raw images are smaller due to their representation using unsigned integers, taking up less space \pagebreak[4] compared to floating point numbers used for encoded samples. The encoded samples have an advantage due to the encoder's pre-training, therefore they nonetheless improve the performance.

\subsection{Ablation: Effect of Compression}
The previous section examined the effect of an encoder in isolation, while this section examines the effect of a compressor in the same way. For the experiment, the number of memory slots was fixed at $N=10000$, while the compression parameter $k$ was varied.

The result can be seen in \cref{fig:fixed_slots}. The autoencoder could only be used for the experiments without the fixed encoder because the spacial dimensions of the encoded featuremaps (2$\times$2) are too small to perform convolutions and pooling. It also becomes clear that the performance of the baseline cannot be reached using this architecture.
Increased compression negatively impacts performance across all compressors, which is expected. Remarkably, the upper bound of the curves reflects the optimal accuracy achievable with this configuration.
The quantization strategy approaches baseline performance, given a sufficiently high compression parameter. The best accuracy is reached between $k=8$ and $k=16$, which corresponds to compression of over 85~\%.
The thinning compression performs significantly better when a fixed encoder is used.

\subsection{Performance on a fixed Memory Budget}
This experiment aims to show how well different configurations perform under a memory constraint. To replicate these conditions, the total memory consumption was fixed at 4~MiB for CIFAR10 and at 6~MiB for CIFAR100.
The memory is filled with $N$ samples up to the maximum available storage size.

\begin{figure}[t]
    \centering
    \includegraphics[width=\linewidth]{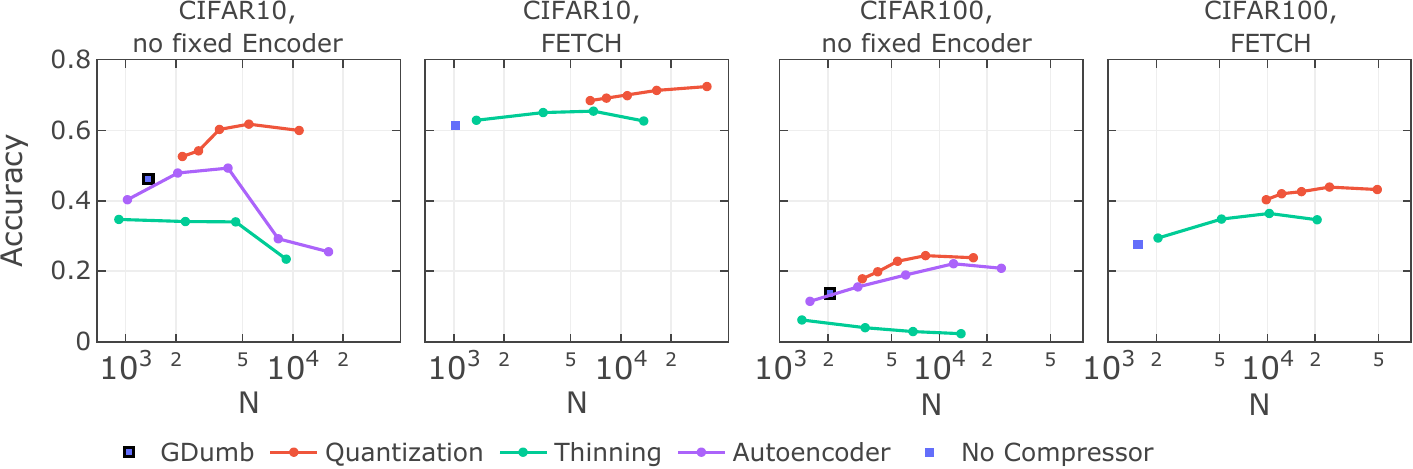}
    \vspace{-6mm}
    \caption{Performance on a fixed memory budget}
    \label{fig:fixed_memory_mb}
    \vspace{-5mm}
\end{figure}

The results can be seen in \cref{fig:fixed_memory_mb}. Notably, almost all curves show a maximum, where the balance between the number and quantity of the samples in memory is optimal. Exceptions include the quantization strategy in the first plot (where compression is always beneficial) and the thinning compressor in the second and fourth plots, where compression is always harmful.
As previously discussed, the results show that encoding improves performance, as evidenced by the higher accuracy of the setups using FETCH.

\subsection{Comparison with other approaches}
We compare FETCH with the following state-of-the-art approaches: \emph{REMIND} \cite{Hayes20}, freezes the early layers of a ResNet and performs product quantization on the resulting featuremaps, before storing them in memory. During continual learning, the data from the continual stream is mixed with samples from the memory. 
\emph{ACAE-REMIND} \cite{Wang21a} extends REMIND with an additional autoencoder to compress the samples even further.
\emph{`Smaller Is Better'}, the best-performing variant of an approach proposed in \cite{Pelosin21}, involves resizing raw images to 8$\times$8 pixels before storing them in the episodic memory. A network is retrained from scratch, whenever the input distribution changes.
To the best of our knowledge, no method besides FETCH combines the benefits of freezing early layers of a convolutional neural network with compressed replay in the pipeline of GDumb.

We evaluate FETCH in three different settings:
\emph{Setting A}: the encoder is pre-trained on TinyImageNet. The classification head is initialized randomly. This approach was used in the other sections of this paper.
\emph{Setting B}: both the encoder and the classification head are pre-trained using the data from the first tasks (in the case of CIFAR10 the classes \texttt{Airplane} and \texttt{Automobile}). This setting is used by REMIND and ACAE-REMIND.
\emph{Setting C}: the encoder and classification head were pre-trained on TinyImageNet.

\begin{figure}[t]
    \centering
    \includegraphics[width=\textwidth]{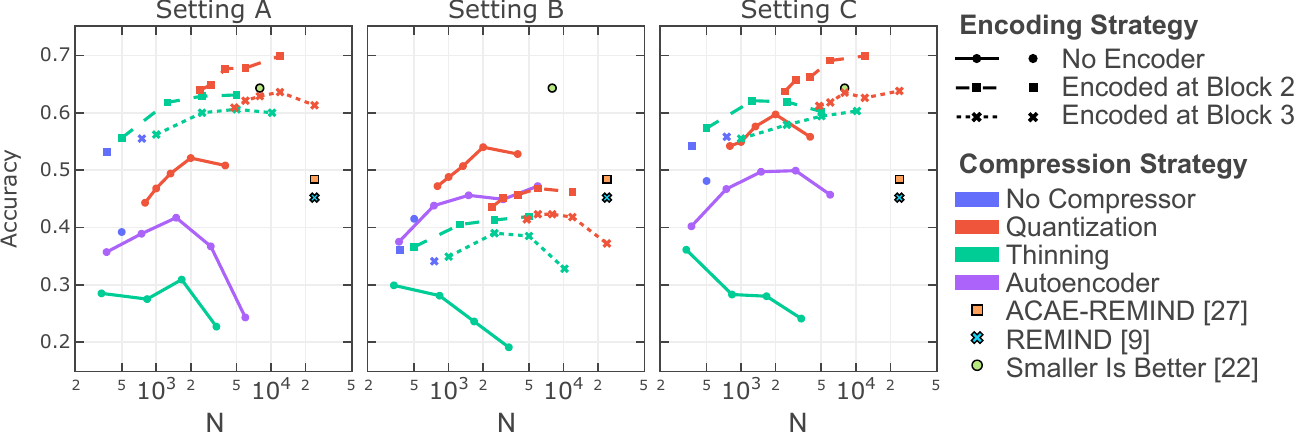}
    \caption{Results for a varying compression parameter $k$ and a fixed total memory of 1.536 MB, following \cite{Wang21a}. The data for `REMIND' and `ACAE-REMIND' is sourced from \cite{Wang21a}; the result for `Smaller Is Better' was produced by the Code provided by \cite{Pelosin21}. Compared approaches are outlined.}
    \label{fig:comparison_to_ohters}
    \vspace{-5mm}
\end{figure}

We vary the compression parameter and the layer up to which we keep the weights frozen. The memory is filled with $N$ samples up to a maximum size of 1.536 MB. All compared methods use ResNet-18 and the CIFAR10 dataset. The results are shown in figure \cref{fig:comparison_to_ohters}.
Setting B shows that the simple quantization strategy performs similarly to REMIND and ACAE-REMIND, even outperforming both approaches for some configurations.
Comparing settings A and B shows the positive influence of the diverse pre-training dataset. Setting A shows that pre-training enables FETCH to also outperform `Simpler Is Better'. Our experiments in setting C show a positive effect of using in-distribution datasets for pre-training the classification head.

\section{Conclusion}
\label{sec:conclusion}
This work aimed at investigating the advantages of compressed replay in the context of GDumb.
We evaluated the effect of different compression strategies as well as the effect of pre-training and freezing parts of the backbone.
A combination of both techniques showed improved performance over both GDumb and selected compressed replay techniques in our experiments.
These findings suggest that episodic memories with a large number of compressed exemplars and the transfer effects of pre-trained components benefit replay learning.
However, FETCH has limitations, including the need to retrain the classification head whenever the data distribution changes.
In future work, we wish to investigate the proposed two-step compression scheme in combination with other memory-based CL approaches and domains outside computer vision.
Although FETCH can directly be used in applications with limited memory, such as mobile robotics, this work intends to serve as a baseline for future research in the area of memory-constrained CL.

\bibliographystyle{splncs04}
\bibliography{literature}
\end{document}